\documentclass[sigconf]{acmart}
\acmSubmissionID{612}

\usepackage{bm}

\usepackage[ruled]{algorithm2e} 
\usepackage{multirow}
\usepackage{enumitem}

\SetAlFnt{\small}
\SetAlCapFnt{\small}
\SetAlCapNameFnt{\small}
\SetAlCapHSkip{0pt}
\usepackage{pifont}
\usepackage[T1]{fontenc}
\newcommand{\cmark}{\ding{51}}

\newcommand{\xmark}{\ding{55}}

\definecolor{ao}{rgb}{0.0, 0.0, 1.0}
\definecolor{airforceblue}{rgb}{0.36, 0.54, 0.66}
\definecolor{ceruleanblue}{rgb}{0.16, 0.32, 0.75}
\definecolor{cerulean}{rgb}{0.0, 0.48, 0.65}
\definecolor{celestialblue}{rgb}{0.29, 0.59, 0.82}
\definecolor{azure(colorwheel)}{rgb}{0.0, 0.5, 1.0}
\definecolor{babyblue}{rgb}{0.54, 0.81, 0.94}
\definecolor{babyblueeyes}{rgb}{0.63, 0.79, 0.95}
\definecolor{ballblue}{rgb}{0.13, 0.67, 0.8}

\definecolor{asparagus}{rgb}{0.53, 0.66, 0.42}
\definecolor{ao(english)}{rgb}{0.0, 0.5, 0.0}
\definecolor{applegreen}{rgb}{0.55, 0.71, 0.0}
\definecolor{armygreen}{rgb}{0.29, 0.33, 0.13}
\definecolor{gray-asparagus}{rgb}{0.27, 0.35, 0.27}
\definecolor{green(ryb)}{rgb}{0.4, 0.69, 0.2}

\definecolor{amethyst}{rgb}{0.6, 0.4, 0.8}
\definecolor{antiquefuchsia}{rgb}{0.57, 0.36, 0.51}
\definecolor{blue-violet}{rgb}{0.54, 0.17, 0.89}
\definecolor{brightlavender}{rgb}{0.75, 0.58, 0.89}
\definecolor{brightube}{rgb}{0.82, 0.62, 0.91}
\definecolor{brilliantlavender}{rgb}{0.96, 0.73, 1.0}

\definecolor{amber}{rgb}{1.0, 0.75, 0.0}
\definecolor{amber(sae/ece)}{rgb}{1.0, 0.49, 0.0}
\definecolor{atomictangerine}{rgb}{1.0, 0.6, 0.4}
\definecolor{burntorange}{rgb}{0.8, 0.33, 0.0}
\definecolor{burntsienna}{rgb}{0.91, 0.45, 0.32}
\definecolor{cadmiumorange}{rgb}{0.93, 0.53, 0.18}
\definecolor{carrotorange}{rgb}{0.93, 0.57, 0.13}
\definecolor{chocolate(web)}{rgb}{0.82, 0.41, 0.12}
\definecolor{chromeyellow}{rgb}{1.0, 0.65, 0.0}
\definecolor{darkorange}{rgb}{1.0, 0.55, 0.0}
\definecolor{darktangerine}{rgb}{1.0, 0.66, 0.07}
\definecolor{deepcarrotorange}{rgb}{0.91, 0.41, 0.17}
\definecolor{deepsaffron}{rgb}{1.0, 0.6, 0.2}
\definecolor{fulvous}{rgb}{0.86, 0.52, 0.0}

\newcommand{\name}{NPVA}
\newcommand{\fullname}{Neural Point-based Volumetric Avatar}

\copyrightyear{2023}
\acmYear{2023}
\setcopyright{rightsretained}
\acmConference[SA Conference Papers '23]{SIGGRAPH Asia 2023 Conference Papers}{December 12--15, 2023}{Sydney, NSW, Australia}
\acmBooktitle{SIGGRAPH Asia 2023 Conference Papers (SA Conference Papers '23), December 12--15, 2023, Sydney, NSW, Australia}
\acmDOI{10.1145/3610548.3618204}
\acmISBN{979-8-4007-0315-7/23/12}

\begin{document}

\title[Neural Point-based Volumetric Avatar]{Neural Point-based Volumetric Avatar: Surface-guided Neural Points for Efficient and Photorealistic Volumetric Head Avatar}

\author{Cong Wang}
\affiliation{%
 \institution{Tsinghua University, Tencent AI Lab}
 \country{China}
}
\email{wangcong20@mails.tsinghua.edu.cn}
\author{Di Kang}
\affiliation{%
 \institution{Tencent AI Lab}
 \country{China}
}
\email{di.kang@outlook.com}
\author{Yan-Pei Cao}
\affiliation{%
 \institution{Tencent AI Lab}
 \country{China}
}
\email{caoyanpei@gmail.com}
\author{Linchao Bao}
\affiliation{%
 \institution{Tencent AI Lab}
 \country{China}
}
\email{linchaobao@gmail.com}
\author{Ying Shan}
\affiliation{%
 \institution{Tencent AI Lab}
 \country{China}
}
\email{yingsshan@tencent.com}
\author{Song-Hai Zhang}
\authornote{Song-Hai Zhang is the corresponding author (shz@tsinghua.edu.cn).}
\affiliation{%
 \institution{Tsinghua University}
 \country{China}
}
\email{shz@tsinghua.edu.cn}

\begin{teaserfigure}
\centering
\includegraphics[width=0.95\textwidth]{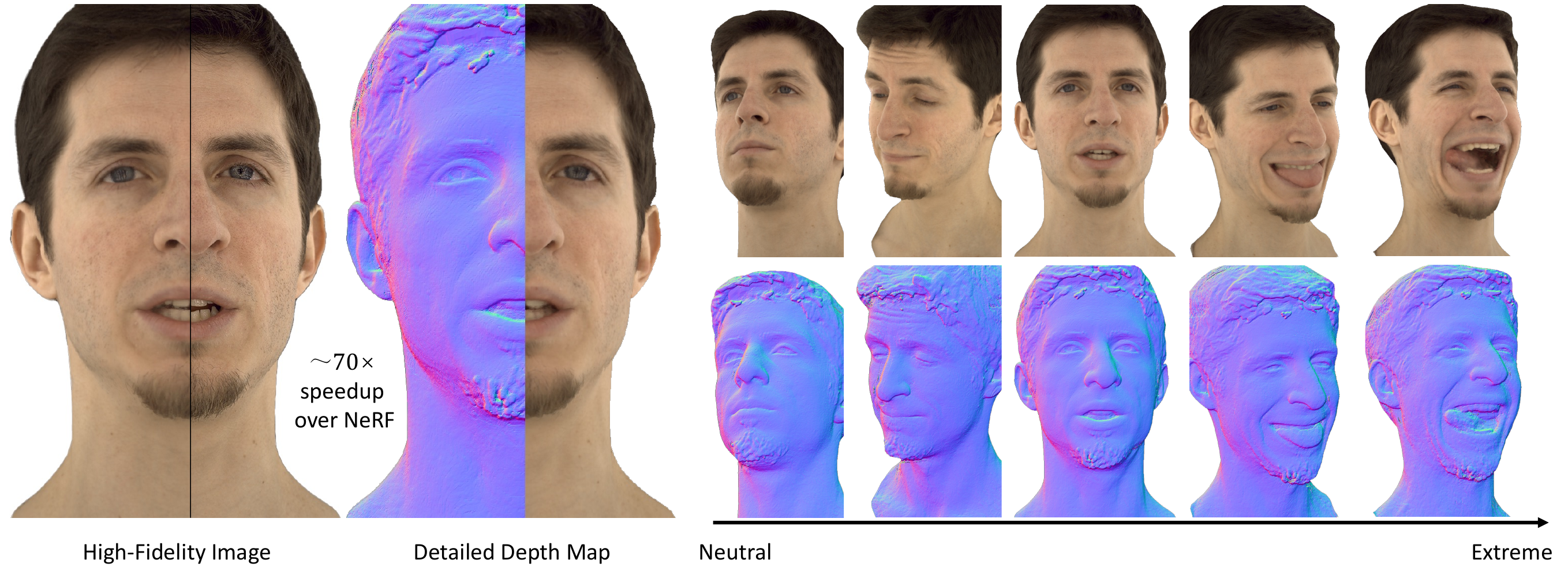}
\caption{
{\fullname} explores point-based neural representation combined with volume rendering, achieving high-fidelity facial animations (images and depth maps) while maintaining efficiency comparable to mesh-based methods (Tab.~\ref{tab:view_synthesis}).
During training, {\name} can adaptively allocate more points to challenging facial regions, forming a thicker ``shell'' (i.e., a higher variance of projected distances onto the face in the normal direction) and increasing capacity as needed. 
In the leftmost image, we show our rendering and the ground truth (GT) side-by-side on the right and left parts, respectively.
Observe the close resemblance between our rendering quality and the GT.
}
\label{fig:teaser}
\end{teaserfigure}

\begin{abstract}
Rendering photorealistic and dynamically moving human heads is crucial for ensuring a pleasant and immersive experience in AR/VR and video conferencing applications.
However, existing methods often struggle to model challenging facial regions (e.g., mouth interior, eyes, and beard), resulting in unrealistic and blurry results.
In this paper, we propose {\fullname} ({\name}), a method that adopts the neural point representation as well as the neural volume rendering process and discards the predefined connectivity and hard correspondence imposed by mesh-based approaches.
Specifically, the neural points are strategically constrained around the surface of the target expression via a high-resolution UV displacement map, achieving increased modeling capacity and more accurate control.
We introduce three technical innovations to improve the rendering and training efficiency: a patch-wise depth-guided (shading point) sampling strategy, a lightweight radiance decoding process, and a Grid-Error-Patch (GEP) ray sampling strategy during training.
By design, our {\name} is better equipped to handle topologically changing regions and thin structures while also ensuring accurate expression control when animating avatars.
Experiments conducted on three subjects from the Multiface dataset demonstrate the effectiveness of our designs, outperforming previous state-of-the-art methods, especially in handling challenging facial regions.
\end{abstract}

%
%


%
%


\maketitle

\section{Introduction}

Realizing photorealistic rendering of an animatable human head is a pivotal goal in computer graphics and vision,
which has broad applications such as AR/VR communications~\cite{DBLP:conf/ismar/HeDP20,DBLP:conf/uist/Orts-EscolanoRF16, DBLP:journals/tog/LombardiSSS18}, gaming~\cite{waggoner2009my}, and remote collaboration~\cite{DBLP:journals/rcim/WangBBZZWHYJ21}.
However, providing a satisfying and immersive experience in these applications remains immensely challenging due to our innate ability to express and perceive emotions through subtle facial cues~\cite{ekman1980face}.
Existing data-driven learning methods often generate noticeable artifacts in the mouth area and blurry beard textures~\cite{DBLP:journals/cgf/ZollhoferTGBBPS18, DBLP:journals/tog/EggerSTWZBBBKRT20, DBLP:conf/eccv/KhakhulinSLZ22, DBLP:conf/cvpr/GrassalPLRNT22, DBLP:journals/tog/LombardiSSS18, DBLP:conf/cvpr/MaSSWLTS21}.
This limitation is primarily attributed to their underlying mesh-based representations, since the predefined mesh has a fixed topology and limited discretization resolution.

To illustrate this limitation, consider the Deep Appearance Model (DAM)~\cite{DBLP:journals/tog/LombardiSSS18} that decodes a head mesh and the corresponding view-specific texture UV map.
Despite achieving high-quality rendering results for the skin regions, DAM produces conspicuous artifacts in the mouth and hair regions due to inaccurate correspondences across frames (e.g., mouth interior) and the mesh's inability to model thin structures (e.g., beard).
To alleviate these issues, Pixel Codec Avatar (PiCA)~\cite{DBLP:conf/cvpr/MaSSWLTS21} proposes the use of neural textures, allowing the subsequent neural renderer to address inaccurate shape estimation and topological inconsistencies (e.g., closed/open mouth), attaining moderately improved facial geometry and renditions for the mouth region.
Similarly, Mixture of Volumetric Primitives (MVP)~\cite{DBLP:journals/tog/LombardiSSZSS21} attaches volumetric primitives (predicted by a CNN) to mesh vertices, replacing vertex colors with more flexible volumetric primitives.
However, these enhanced ``texture''-like representations remain embedded in a predefined topology and tend to produce blurry results if inaccurate correspondences occur.

Therefore, we propose {\fullname} ({\name}) that leverages highly flexible neural points~\cite{DBLP:conf/eccv/AlievSKUL20} and versatile neural volume rendering, enabling sharper rendering for both topologically changing geometries (e.g., mouth interior) and translucent thin beard structures.

To create animatable head avatars, another key technical challenge lies in enhancing the controllability of neural points to generate accurate target expressions.
To address this, we reconstruct an intermediate coarse geometry (represented as a UV position map) of the driving signal (i.e., target expression) and constrain the movable neural points close to its surface.
Further, to maximize the benefits of neural volume rendering, we introduce an additional displacement map, which allows the points to move to more optimal positions around the surface.
For instance, after training, more points are located inside the mouth, resulting in a thicker ``point shell'' and increased modeling capability for volume rendering.

Efficient rendering and training are also indispensable for practical applications.
To this end, we propose three technical innovations.
(1) We introduce a novel depth-guided~\cite{DBLP:conf/siggrapha/LinPXYSBZ22} sampling method that incorporates local depth context information (i.e. a patch), achieving more realistic rendering while reducing the rendering time by $\sim 10 \times$ compared to the vanilla NeRF.
(2) We develop a lightweight radiance decoding process that eliminates unnecessary per-point processing used in~\cite{DBLP:conf/cvpr/XuXPBSSN22, DBLP:conf/cvpr/BaiTHSTQMDDOPTB23}, significantly improving rendering efficiency ($\sim 7 \times$) and offering better generalization in our dynamic modeling task.
(3) Lastly, to speed up training, we propose a novel Grid-Error-Patch (GEP) ray sampling strategy comprising three stages: a uniform grid-sampling stage for rapid initialization of a coarse result, an error-based importance sampling that delves into more challenging regions, and a patch-based stage to impose high-level perceptual image losses.

Our contributions are summarized below:
\begin{itemize}
    \item We propose a novel volumetric representation based on neural points that are dynamically allocated around the surface (i.e., the target expression) for animatable head avatar creation.
    This representation is inherently capable of better handling thin geometry and topological changes.
    \item We introduce three technical innovations to enable efficient rendering and training, including 
    a patch-wise depth-guided shading point sampling method, 
    a lightweight radiance decoding process, 
    and a Grid-Error-Patch training strategy.
    \item Experiments on the Multiface dataset show that our approach produces higher-quality images for novel expressions and novel views while being $\sim 70 \times$ faster than NeRF.
\end{itemize}

\begin{figure*}[tb]
\centering
\includegraphics[width=0.95\textwidth]{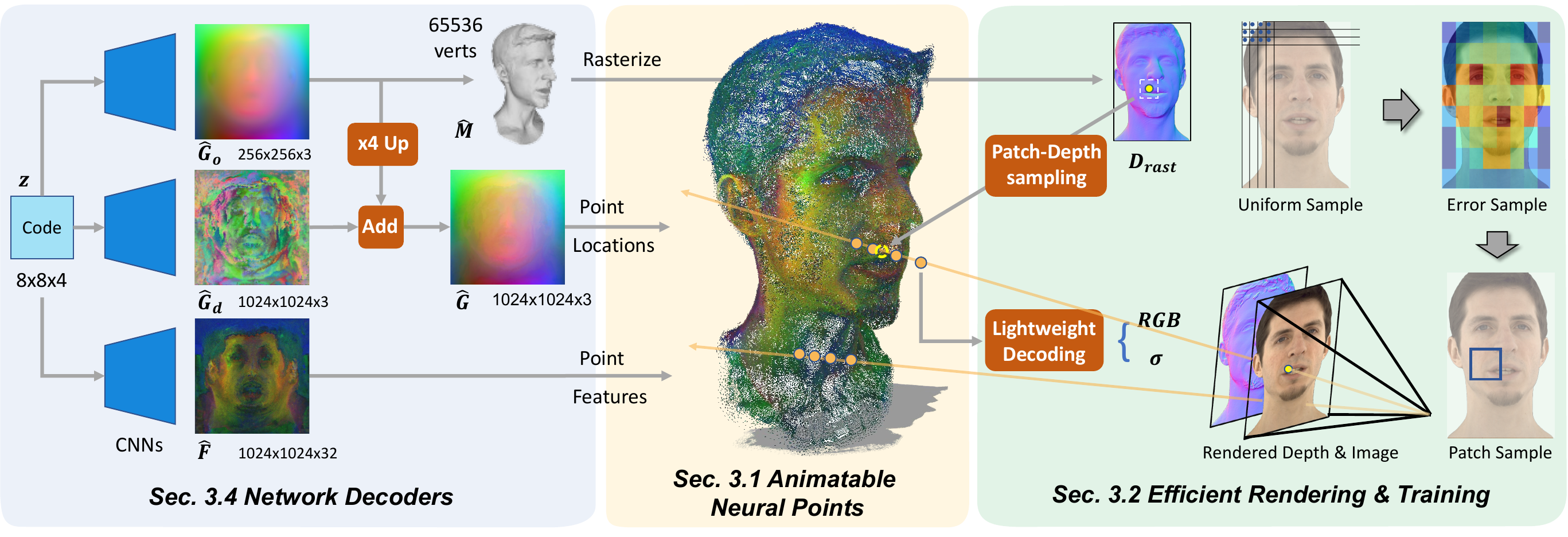}
\caption{\textbf{Overview of {\fullname} ({\name}).}
The core of our approach is a neural point-based volumetric representation (middle),
with points distributed around the surface of the target expression. This surface is defined by the low-resolution position map $\hat{\bm{G}}_o$ with intermediate supervision.
A high-resolution displacement map $\hat{\bm{G}}_d$ allows the points to adaptively move within a certain range, as needed to provide increased capacity in more challenging regions (e.g., mouth, hair/beard).
The attached point features are obtained from the feature map $\hat{\bm{F}}$.
$\hat{\bm{G}}_o$, $\hat{\bm{G}}_d$, and $\hat{\bm{F}}$ are decoded from the latent code $\bm{z}$ (left), which is trained in a variational auto-encoding style (encoder omitted).
In addition, we propose three technical innovations with the aim of achieving rendering efficiency on par with mesh-based methods (right).
}
\label{fig: pipeline}
\end{figure*}

\section{Related Works}

\subsection{Continuous Neural Representations.}
Neural implicit representations, such as ONet~\cite{DBLP:conf/cvpr/MeschederONNG19}, DeepSDF~\cite{DBLP:conf/cvpr/ParkFSNL19}, and NeRF~\cite{DBLP:conf/eccv/MildenhallSTBRN20}, parameterize a continuous occupancy/SDF/radiance fields with a network (e.g., a multilayer perceptrons, MLP).
The property (e.g., occupancy, color/density) of any location in the scene could be queried by feeding its coordinate (optionally along with specified viewing direction) to the MLP, which is also referred as the coordinate network.

Rapid progresses have been achieved on shape modeling~\cite{DBLP:conf/cvpr/ChenZ19, DBLP:conf/nips/WangLLTKW21, DBLP:journals/tog/MartelLLCMW21, DBLP:conf/eccv/TretschkTGZST20, DBLP:conf/cvpr/WuYLCZGLZ23} and appearance modeling~\cite{DBLP:conf/nips/SitzmannZW19, DBLP:conf/nips/SchwarzLN020, DBLP:conf/cvpr/KellnhoferJJSPW21, DBLP:conf/cvpr/ZhengZY023}.
For example, PVA~\cite{DBLP:journals/corr/abs-2101-02697} and KeypointNeRF~\cite{DBLP:conf/eccv/MihajlovicBZ0S22} obtain facial neural radiance fields from multi-view images for high-fidelity static face reconstruction.
Later, these neural implicit representations are further extended to handle dynamic scenes (e.g., a talking head) by introducing deformation techniques~\cite{yu2023nofa,DBLP:conf/cvpr/GafniTZN21,DBLP:journals/corr/abs-2201-00791,tang2022explicitly,wu2023speech2lip}.
For instance, NerFace~\cite{DBLP:conf/cvpr/GafniTZN21} conditions the NeRF on 3DMM coefficients, making a small step to animatable avatar creation.
DFA-NeRF~\cite{DBLP:journals/corr/abs-2201-00791} further conditions NeRF with disentangled face attribute features and enables more detailed control over the talking head.

Methods using neural implicit representations assume no fixed topology and have infinite resolution (in theory), achieving impressive novel view synthesis results for dynamic scenes.
However, they usually suffer from various artifacts (e.g., blur, distortion) and inaccurate control when rendering novel expressions/poses, which is critical for creating high-quality animatable avatars.
In contrast, our approach employs an explicit point-based representation driven by a coarse surface (represented as a UV position map) of the novel expression, achieving better generalization on novel expression and precise expression control.

\subsection{Discrete Neural Representations.}
Common discrete representations in CV/CG include mesh, grid/voxel, and point cloud.
Recently, these representations have been extended with neural features to not only reconstruct the shape, but also create photorealistic image renderings thanks to the rapid progress in differentiable neural rendering~\cite{DBLP:conf/cvpr/RakhimovALB22, DBLP:journals/tog/LaineHKSLA20, DBLP:journals/tog/RuckertFS22, DBLP:conf/cvpr/LassnerZ21, DBLP:journals/tog/CaoSKSZSLWBYSS22, DBLP:conf/cvpr/MaZQLZ23, DBLP:conf/icra/WangWM22, DBLP:conf/iros/WangZWDQM21}.

For example, neural volumes~\cite{DBLP:journals/tog/LombardiSSSLS19} learns a radiance field in the canonical space and introduce another warp field to handle dynamic scenes.
However, regular grid/voxel-based representations usually suffer from the required cubic memory footprint and slow rendering speed due to the processing of empty voxels.

Another line of works uses mesh-based neural representations~\cite{DBLP:journals/tog/LombardiSSZSS21,DBLP:conf/cvpr/MaSSWLTS21, DBLP:conf/cvpr/WangKSQWBZ25, DBLP:conf/iccv/WangWM23}.
For example,
PiCA employs a neural UV texture map to render the head avatar in different expressions from any given viewpoint.
Neural head avatar~\cite{DBLP:conf/cvpr/GrassalPLRNT22} introduces additional geometry and texture networks in complementary to a base FLAME~\cite{DBLP:journals/tog/LiBBL017} head model, resulting in a better shape and texture modeling.
However, mesh-based representations require accurate surface geometry and semantically consistent correspondence, which is usually hard to acquire.
Alternatively, MVP combines mesh and grid-based representations, and uses the volume rendering technique to render images.
However, the color and density of MVP's primitives are decoded from 2D CNNs and further aggregated to obtain the radiance of query points, which inevitably causes blurry renderings.
As a result, we propose to explore a more flexible neural point-based representation.
Concurrent with our work, PointAvatar~\cite{DBLP:conf/cvpr/ZhengYWBH23} also uses a point-based representation.
Different from ours, their points only store color information and use splatting for rendering.
We utilize powerful volume rendering, which has the potential to render higher-quality hairs and beards.

\subsection{Differentiable Neural Rendering.}
Rapid progress in differentiable neural rendering~\cite{DBLP:conf/cvpr/0002JHZ20,DBLP:conf/iccv/Liu0LL19,DBLP:conf/cvpr/LiuZPSPC20, DBLP:journals/tog/MullerESK22, DBLP:conf/siggrapha/WangLBXZW19, DBLP:journals/corr/abs-2508-09597} makes analysis-by-synthesis pipeline more powerful than before and enables avatar creation from even a short monocular video~\cite{DBLP:conf/cvpr/ZhengYWBH23,DBLP:conf/cvpr/GrassalPLRNT22}.
For example, deferred neural rendering~\cite{DBLP:journals/tog/ThiesZN19} proposes a neural texture map, which is decoded by a neural renderer into high quality renderings from any viewpoint.
Recently, neural volume rendering~\cite{DBLP:conf/eccv/MildenhallSTBRN20} achieves impressive high-quality renderings by introducing volume rendering to neural implicit fields.
What's more, volume rendering can naturally model translucent objects (e.g., hair, smoke) in very high fidelity, which is very suitable for head modeling.
However, a major drawback of neural volume rendering is low rendering efficiency.
Our approach employs volume rendering to ensure high-fidelity visual results, and proposes a series of acceleration methods to achieve rendering speed matching neural texture rendering.

\section{Methods}

An overview of our {\fullname} ({\name}) is shown in Fig.~\ref{fig: pipeline}.
Given a latent code corresponding to the target facial state, we employ three decoders to generate a position map $\hat{\bm{G}}_o$, a displacement map $\hat{\bm{G}}_d$, and a feature map $\hat{\bm{F}}$, respectively.
Our {\name} representation is constructed from these maps, followed by a point-based neural volume rendering to efficiently produce high-fidelity images and detailed depth maps from any viewpoint.

In Sec.~\ref{sec:representation}, we introduce our {\name} representation that leverages flexible point clouds for improved modeling of topological changes and thin structures, and utilizes volume rendering to produce high-fidelity images.
In Sec.~\ref{sec:efficient}, we present our efficient rendering and training strategies, enabling {\name} to render photorealistic images comparable to NeRF while being $\sim 70 \times$ faster.
Sec.~\ref{sec:losses} details our training losses. 
Sec.~\ref{sec:net} describes our implementation details.

\subsection{\fullname} 
\label{sec:representation}

\subsubsection{Neural Radiance Field}
\label{sec:nerf}

NeRF~\cite{DBLP:conf/eccv/MildenhallSTBRN20} effectively encodes a static scene using a Multi-Layer Perceptron network (MLP), achieving unprecedented high-quality novel view synthesis results.
During rendering, the trained MLP takes the scene coordinates $\bm{x} \in \mathbb{R}^3$ and viewing direction $(\theta, \phi)$ as input and produces the corresponding density $\sigma_x$ and view-dependent color $\bm{c}_{\bm{x}}$.
The final color of an image pixel is obtained via volume rendering (Eq.~\eqref{eq:vr}), which integrates all shading points on the ray that passes through this image pixel.
Mathematically, NeRF uses piece-wise constant density and color as an approximation:
\begin{equation}
\setlength{\abovedisplayskip}{3pt}  
\setlength{\belowdisplayskip}{3pt}  
c = \sum^N_{i=1} T_i (1 - \exp (- \sigma_i \delta_i)) c_i, 
\;\;T_i = \exp(-\sum^{i-1}_{j=1} \sigma_j \delta_j)
\label{eq:vr}
\end{equation}
where $\sigma_i$ and $\bm{c}_i$ denote density and color, respectively,  
and $\delta_i$ is the distance between two adjacent shading points.

\subsubsection{Animatable Neural Points}
\label{sec:neural-points}

We employ \emph{explicit} neural points in our {\name} to achieve more controllable deformation of the underlying implicit neural radiance field.
Leveraging this geometry proxy can also significantly enhance the rendering efficiency (Sec.~\ref{sec:efficient}).

Inspired by Point-NeRF~\cite{DBLP:conf/cvpr/XuXPBSSN22}, our representation consists of a set of neural points, denoted as $\mathcal{A} = \{(\bm{p}_i, \bm{f}_i) | i=1, \ldots, N\}$, where $\bm{p}_i \in \mathbb{R}^3$ denotes the location of point $i$, and $\bm{f}_i$ represents its associated feature. 
In our network, these features are learned in a variational auto-encoding style, similar to DAM~\cite{DBLP:journals/tog/LombardiSSS18} and PiCA~\cite{DBLP:conf/cvpr/MaSSWLTS21}, to create an animatable avatar.
Specifically, the geometry sub-network in the decoder predicts a UV position map $\hat{\bm{G}}_o$ and a UV displacement map $\hat{\bm{G}}_d$.
$\hat{\bm{G}}_o$ stores the vertex positions of a coarse head mesh. We apply additional intermediate supervision on the $256^2$ position map $\hat{\bm{G}}_o$ to obtain better expression control.
To achieve better expressiveness of the point-based neural radiance field, we upsample the position map $\hat{\bm{G}}_o$ to $1024^2$ and incorporate the high-resolution displacement map $\hat{\bm{G}}_d$ to compensate for inaccuracy contained in the coarse geometry. By compositing the upsampled $\hat{\bm{G}}_o$ and $\hat{\bm{G}}_d$, we determine the positions of the neural points. 
The final neural points can adjust their positions adaptively around the surface, as we only apply a regularization term to penalize unreasonably large displacements.

\begin{figure}[tb]
\centering
\includegraphics[width=0.95\linewidth]{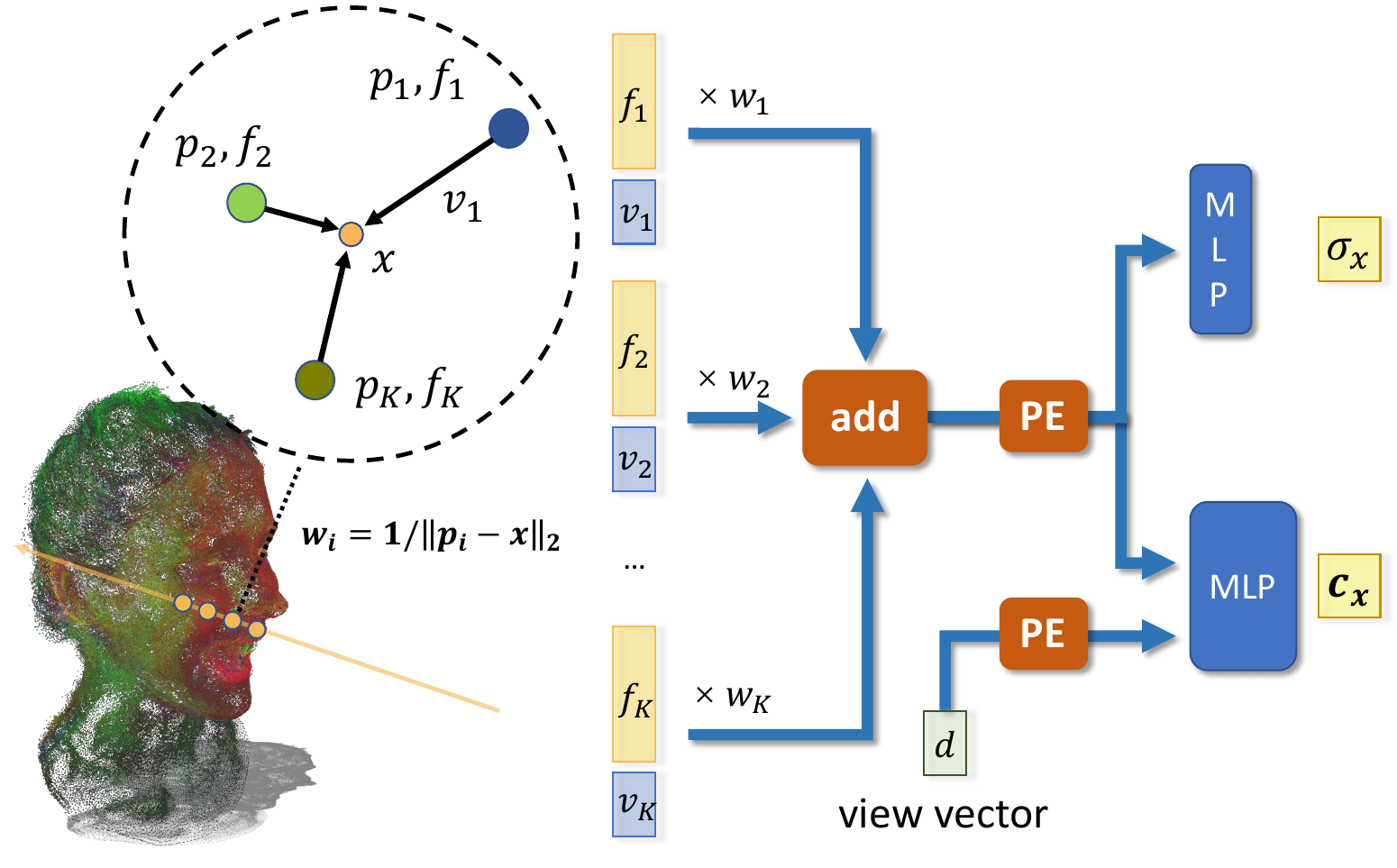}
\caption{\textbf{Lightweight Radiance Decoding.} 
Given a query point, we find its $K$ nearest neighboring neural points.
We weighted sum these points to obtain an ``average'' feature for the subsequent radiance decoding.
Note that we removed the per-point processing MLP used in Point-NeRF.
This lightweight decoding process runs faster and obtains better generalization on novel expressions for our dynamic modeling task.
}
\label{fig:radiance_decoding}
\end{figure}

\subsubsection{Lightweight Radiance Decoding}
\label{sec: NVPA Decoding}

The radiance (i.e., color $\bm{c}_{\bm{x}}$ and density $\sigma_{\bm{x}}$) at position $\bm{x} \in \mathbb{R}^3 $ is extracted based on (up to) its $ K $ nearest neighboring points and a neural decoding MLP (see Fig.~\ref{fig:radiance_decoding}), inspired by prior methods~\cite{DBLP:conf/cvpr/XuXPBSSN22,DBLP:journals/corr/abs-2206-01290}.
To increase efficiency and ensure better generalization on novel expressions, we design a novel lightweight radiance decoding process that directly aggregates the neural points and passes this ``average'' feature to a lightweight neural decoding network, achieving $\sim 7 \times$ speedup (compared to Point-NeRF) and better renderings.
Specifically, for a given shading point, we compute a weighted average of the features and relative positions of (up to) its $K$ nearest neural points inside a sphere of radius $R$ as in Point-NeRF:
\begin{equation}
\setlength{\abovedisplayskip}{3pt}  
\setlength{\belowdisplayskip}{3pt}  
    \bm{f_x} = \sum_i \frac{w_i}{\sum w_i} \bm{f}_i, \;\; \bm{v_x} = \sum_i \frac{w_i}{\sum w_i} \bm{v}_i
    \label{eq:aggregation}
\end{equation}
where $\bm{v}_i \in \mathbb{R}^6$ is the position encoding of the displacement vector between neural point $\bm{p}_i$ and the shading point $\bm{x}$,
$w_i = \frac{1}{|| \bm{p_i} - \bm{x} ||_2}$ is the weight that is inversely proportional to the Euclidean distance between $\bm{x}$ and $\bm{p}_i$.
Then, two shallow MLPs are applied to decode this average feature to density $\sigma_{\bm{x}}$ and view-dependent color $\bm{c_x}$:
\begin{equation}
\setlength{\abovedisplayskip}{3pt}  
\setlength{\belowdisplayskip}{3pt}  
\mathcal{F}_d: (\bm{f_x}, \bm{v_x}) \rightarrow \sigma_x, \;\; \mathcal{F}_c: (\bm{f_x}, \bm{v_x}, \bm{d}) \rightarrow \bm{c_x}
\end{equation}
Positional encoding~\cite{DBLP:conf/eccv/MildenhallSTBRN20} is applied to every dimension of the input vector in $\mathcal{F}_d$ and $\mathcal{F}_c$.
Finally, a pixel color is obtained via the integration in Eq.~\eqref{eq:vr}.

The primary difference between our radiance decoding and previous methods~\cite{DBLP:conf/cvpr/XuXPBSSN22, DBLP:conf/cvpr/BaiTHSTQMDDOPTB23} is that we omit the per-point feature processing MLP before the feature aggregation in Eq.~\eqref{eq:aggregation}.
This modification provides two benefits: it results in $\sim 7\times$ speedup and improved generalization to unseen expressions in our dynamic modeling task (Fig.~\ref{fig:decoding_comp}).

\begin{figure*}[tb]
\centering
\includegraphics[width=0.95\textwidth]{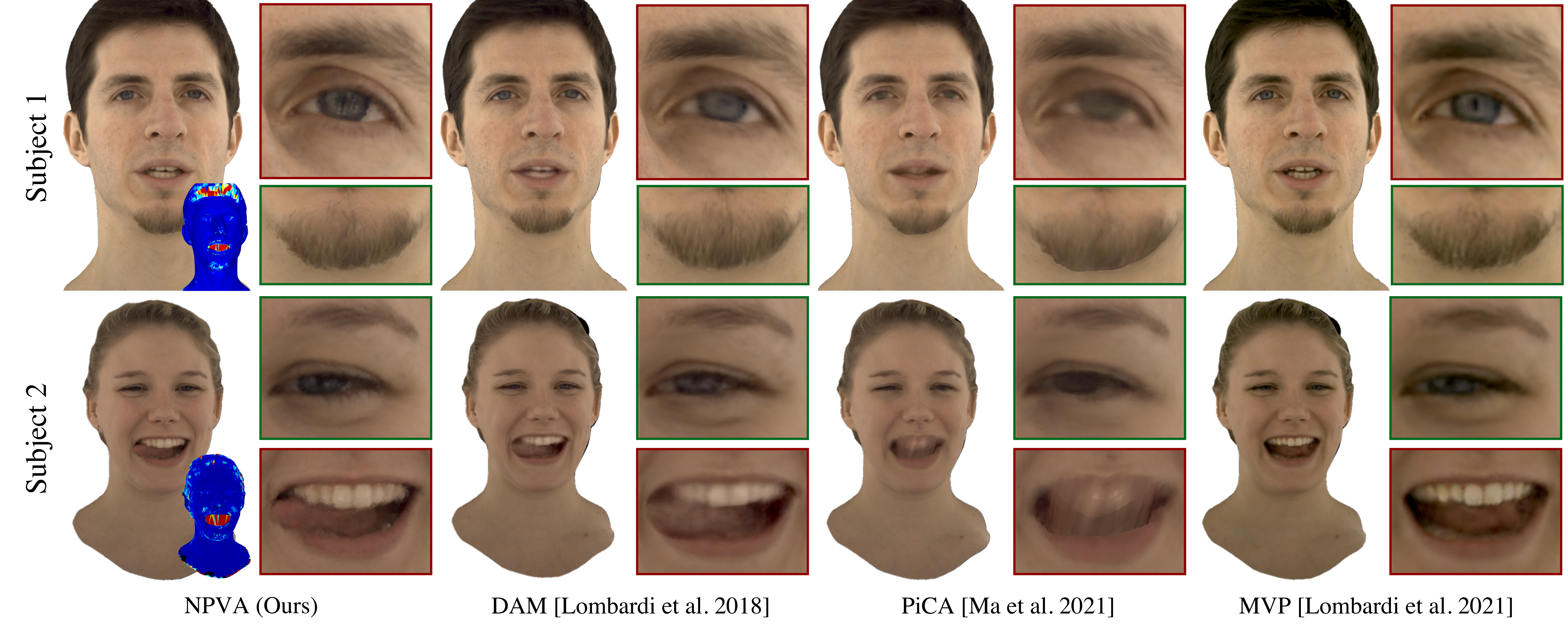}
\caption{\textbf{Qualitative Comparisons with State-of-the-Art Methods.} 
Our NPVA produces more photorealistic facial renditions on held-out test expressions compared to previous state-of-the-art methods, particularly in challenging facial regions (i.e., eyes, beard, and mouth interior).
The normal expression is presented in Row 1, while the extreme expressions are shown in Row 2.
The bottom left corners of the leftmost images show the shell thickness of our {\name}.
The thickness for a specified face is the variance of point-to-surface distance for the points corresponding to it.
Red (blue) indicates larger (smaller) variance.
During learning, our NPVA automatically increases capacity (i.e., thicker shell) to better model the more challenging facial regions (e.g., hair and mouth interior).
}
\label{fig:view_synthesis}
\end{figure*}

\subsection{Efficient Rendering and Training}
\label{sec:efficient}

\subsubsection{Patch-wise Depth-guided Volume Rendering}

Since we have some prior knowledge about the scene (i.e., a head and its coarse shape), we can focus on sampling the shading points around the surface, significantly improving the rendering efficiency compared to the original NeRF.

We propose a patch-wise depth-guided sampling strategy, taking into account that the envelope of a head mesh is often not very accurate and the visible facial parts could appear on different depth levels (e.g., jaw and neck).
Specifically, we define a fixed connectivity on the $256^2$ position map $\hat{\bm{G}}_o$ so that we can easily rasterize a depth map $\bm{D}_\text{rast}$.
For a ray passing through $[p_x, p_y]$, we consider a local depth patch centered on it and obtain the minimum and maximum depth values $D_\text{min}$ and $D_\text{max}$.
In our implementation, we consider only nine pixels, i.e., $\{ \bm{D}_\text{rast} (p_x + i, p_y + j) \;|\; i, j \in \{-s, 0, s\} \}$, where $s$ is a hyper-parameter set to 16 on $1024\times667$ images.

If $D_\text{max} - D_\text{min} < \delta_d$ (smaller than a threshold), there is only one depth level (e.g., no sudden depth changes like jaw and neck), and we use $d_c = (D_\text{max} + D_\text{min}) / 2$ as our sampling center.
If $D_\text{max} - D_\text{min} \ge \delta_d$, there are likely two depth levels, and we split our budget equally between them and sample shading points around $D_\text{min}$ and $D_\text{max}$ separately.
For every depth level $d_c$, we uniformly randomly sample points $\bm{p}(t_i)$ within evenly spaced bins $t_i$ centered at $d_c$ as follows:
\begin{equation}
\setlength{\abovedisplayskip}{3pt}  
\setlength{\belowdisplayskip}{3pt}  
    \bm{p}(t_i) = \bm{o} + t_i \overrightarrow{\bm{d}}, \;\;
    t_i \sim \mathcal{U}[d_c + (\frac{2(i-1)}{N} - 1) r, d_c + (\frac{2 i}{N} - 1) r]
\end{equation}
where $\bm{o}$ is the optical center, 
$\overrightarrow{\bm{d}}$ indicates the view direction, 
$\mathcal{U}$ represents uniform distribution, 
$r$ is the sampling radius and equals 20 in our experiments,
and $N$ is the number of sampling points.

\paragraph{Discussion:} E-NeRF~\cite{DBLP:conf/siggrapha/LinPXYSBZ22} proposes a similar \emph{pixel-wise} depth-guided sampling method, reducing the required shading points for volume rendering.
However, this sampling method only considers the depth of the current pixel and cannot properly handle facial parts that appear on different depth levels (e.g., beard), yielding suboptimal results like mesh-based methods (see Fig.~\ref{fig:d_sampling_comp}).

\subsubsection{GEP Training Strategy}
\label{sec:GEP}

For head image rendering, artifacts usually appear on several difficult but small regions (i.e., mouth, eyes); therefore, uniformly sampling rays covering the entire head region is inefficient.
To address this issue, we propose a three-stage ray sampling strategy that consists of a \underline{G}rid-based uniform sampling stage to initiate the training, an \underline{E}rror-based importance sampling stage to refine the challenging regions, 
and a \underline{P}atch-based sampling stage to improve perceptual quality.

\paragraph{Grid-Sample Stage (G-Stage)}
In this stage, we prioritize full coverage of the images. 
The image is split into equal-sized grids without overlap, and we randomly sample one ray per grid
G-Stage ensures uniform sampling across all regions and generates an initial model that produces reasonable results for all regions of the image.
Moreover, we keep track of the error for each grid, obtaining an error map $\bm{E}$ ($8\times8$) for later error-based importance sampling.

\paragraph{Error-Sample Stage (E-Stage)}
During this E-Stage, we adjust the sampling probabilities of the grids based on the grid error map initialized in the previous G-Stage.
Specifically, the sampling probability of a grid region is proportional to its grid error, resulting in an error-based importance sampling similar to~\cite{zhu2022semi, DBLP:conf/iccv/SucarLOD21}.
As a result, we allocate more computing budget to difficult facial regions (e.g., mouth interior, hair, and eyes in "Error Sample" of Fig.~\ref{fig: pipeline}), significantly improving image quality within the same number of training epochs (see Fig.\ref{fig:training_search_r}a).
Note that we maintain an error map of a smaller size in this stage and dynamically update the sampling probability.

\paragraph{Patch-Sample Stage (P-Stage)}
In this stage, we sample rays belonging to an image patch (instead of individual pixels) so that we can apply patch-based perceptual loss~\cite{DBLP:conf/cvpr/ZhangIESW18}.
Using a perceptual loss along with per-pixel losses (e.g., $L_2$) can help reduce image blur, resulting in sharper images and visually better results~\cite{DBLP:conf/cvpr/MenonDHRR20,DBLP:conf/cvpr/LuoNKXWWH021}.

\begin{table*}[!t]
\centering
\resizebox{0.85\linewidth}{!}{%
\begin{tabular}{l|c|c|c|c|c|c|c|c|c|c|r}
\toprule
\multirow{2}{*}{Methods} & \multicolumn{2}{c|}{Subject 1} & \multicolumn{2}{c|}{Subject 2} & \multicolumn{2}{c|}{Subject 3} & \multicolumn{2}{c|}{Subject 4} & \multicolumn{2}{c|}{Subject 5} & Inference \\
& MSE $\downarrow$ & LPIPS $\downarrow$ & MSE $\downarrow$ & LPIPS $\downarrow$& MSE $\downarrow$ & LPIPS $\downarrow$ & MSE $\downarrow$ & LPIPS $\downarrow$ & MSE $\downarrow$ & LPIPS $\downarrow$ & Time (ms) \\ 
\midrule

PiCA
 & 34.50 & 0.232 & 28.83 & 0.108 & 25.49 & \underline{0.239} & 30.22 & {0.285} & {21.32} & {0.224} & \textbf{73} \\
DAM
 & \underline{28.40} & {\underline{0.208}} & \underline{23.53} & {\underline{0.088}} & \underline{23.18} & {\textbf{0.183}} & {\underline{24.38}} &  {\underline{0.248}} & {\underline{20.21}} & {\underline{0.185}} & {\underline{107}} \\
MVP & 48.59 & {0.242} & 28.25 & {0.102} & 38.82 & {0.262} & {36.23} & {0.268} & {25.46} & {0.230} & {144} \\
{\name} (Ours) & \textbf{23.70} & {\textbf{0.160}} & \textbf{18.16} & {\textbf{0.075}} & \textbf{21.95} & {\textbf{0.183}} & {\textbf{21.88}} & {\textbf{0.180}} & {\textbf{17.13}} & {\textbf{0.141}} & 482 \\ 
\bottomrule
\end{tabular}
}
\caption{\textbf{Comparisons with State-of-the-Art Methods.}
Compared to state-of-the-art methods, {\name} achieves great improvements on LPIPS and MSE (up to 5.37 lower than the $2^\text{nd}$ best) with slightly slower rendering speed.
We bold (underline) the best ($2^\text{nd}$ best) results.
}
\vspace{-2mm}
\label{tab:view_synthesis}
\end{table*}

\begin{table*}[!t]
\centering
\resizebox{0.92\linewidth}{!}{%
\begin{tabular}{c|l|c|c|c|c|c|c|r|r}
\toprule
\multirow{2}{*}{Label} & \multirow{2}{*}{Name} & \multirow{2}{*}{Point Num.} & \multirow{2}{*}{Disp. Map} & KNN Search & Radiance & GEP & Sampling & \multirow{2}{*}{MSE $\downarrow$} & Inference \\
& & & & Radius - R (mm) & Decoding & strategy & Method & & Time (ms) \\
\midrule
 & {\name}-full (Ours) & 1024$\times$1024 & {\cmark} & R=3 & lightweight & {\cmark} & Patch-Depth & 23.70 & 482 \\ 
 \midrule
(a.1) & {\name}-256 & 256$\times$256 & {\xmark} & R=3 & lightweight & {\cmark} & Patch-Depth & 46.45 & 397 \\
(a.2) & {\name}-1k & 1024$\times$1024 & {\xmark} & R=3 & lightweight & {\cmark} & Patch-Depth & 26.36 & 423 \\
(a.3) & {\name}-2k & 2048$\times$2048 & {\xmark} & R=3 & lightweight & {\cmark} & Patch-Depth & 27.36 & 506 \\
\midrule
(b) & {\name}-heavy & 1024$\times$1024 & {\cmark} & R=3 & Point-NeRF & {\cmark} & Patch-Depth & 24.51 & 3129 \\
\midrule
(c.1) & {\name}-noDepth & 1024$\times$1024 & {\cmark} & R=3 & lightweight & {\cmark} & NeRF & 406.49 & 406  \\
(c.2) & {\name}-PixDepth & 1024$\times$1024 & {\cmark} & R=3 & lightweight & {\cmark} & Pixel-Depth & 24.92 & 413 \\
\midrule
(d) & {\name}-R4 & 1024$\times$1024 & \cmark & R=4 & lightweight & \cmark & Patch-Depth & 23.54 & 563 \\
\midrule
(e) & {\name}-noGEP & 1024$\times$1024 & {\cmark} & R=3 & lightweight & {\xmark} & Patch-Depth & 30.08 & 472 \\
%
%
%
\bottomrule
\end{tabular}
}
\caption{
\textbf{Ablation Studies.}
We demonstrate the impact of different components on the results of our NPVA method (evaluated using MSE).
The first row shows our full method as reference.
%
(a) shows the importance of using an extra displacement map, which cannot be replaced by simply increasing point numbers.
%
(b) shows that our lightweight radiance decoding can produce better rendering results on held-out expressions with a faster rendering speed ($\sim 7 \times$).
(c) shows the superiority of our patch-wise depth-guided sampling method.
%
(d) shows that our GEP ray sampling strategy achieves great performance improvements by properly allocating the computation budget and introducing perceptual loss. 
%
}
\vspace{-2mm}
\label{tab:ablation-study}
\end{table*}

\subsection{Training Losses}
\label{sec:losses}

We have a set of images $\{ I^{(i)}\}, i \in \{1, 2, ..., N\}$, 
their corresponding tracked meshes $\{ \mathcal{M}^{(i)}\}$
, UV position maps $\{ \bm{G}^{(i)}_o \}$ converted from $\{ \mathcal{M}^{(i)}\}$, and reconstructed depth maps $\{ \bm{D}^{(i)} \}$ using Metashape software~\cite{metashape}.
Our training losses include 
per-pixel photometric loss $\mathcal{L}_\text{pho}$, 
patch-based perceptual loss $\mathcal{L}_\text{per}$~\cite{DBLP:conf/cvpr/ZhangIESW18},
coarse mesh loss $\mathcal{L}_m$,
two depth losses $\mathcal{L}_d$ and $\mathcal{L}_\text{rd}$,
and three regularization losses $\mathcal{L}_{s}$, $\mathcal{L}_\text{disp}$ and $\mathcal{L}_\text{kl}$.

The appearance losses are defined as:
\begin{equation}
\setlength{\abovedisplayskip}{3pt}  
\setlength{\belowdisplayskip}{3pt}  
\mathcal{L}_{pho} = \sum_{p \in \mathcal{P}} || I^{(i)}_p - \hat{I}^{(i)}_p ||_2, \;\;
\mathcal{L}_{per} = \text{LPIPS} (I^{(i)}_\mathcal{P}, \hat{I}^{(i)}_\mathcal{P})
\end{equation}
where, $\hat{I}$ is our rendering image, and $\mathcal{P}$ is a set of image coordinates used to train our networks.
Note that $\mathcal{L}_\text{per}$ is only used in the last P-stage due to its patch-based property.

The geometric losses, which help generate more controllable head avatars, are defined as:
\begin{equation}
\setlength{\abovedisplayskip}{3pt}  
\setlength{\belowdisplayskip}{3pt}  
\begin{split}
    &\mathcal{L}_m = || G_o - \hat{G}_o ||_2, \\
    &\mathcal{L}_\text{rd} = || (D^{(i)} - {D_\text{rast}}^{(i)}) \odot M_{D_\text{rast}} ||_1, \\
    &\mathcal{L}_d = \sum_{p \in \mathcal{P}} || (D^{(i)}_p - \hat{D}^{(i)}_p) \odot M_D ||_1
\end{split}
\end{equation}
where $\hat{G}_o$ is the decoded $256^2$ position map,
and $D_\text{rast}$ is a coarse depth map rasterized using $\hat{G}_o$,
$\hat{D}$ is our fine depth map obtained using volume rendering.
Note that depth masks $M_D$ and $M_{D_\text{rast}}$ are used to penalize only those pixels whose depth errors are less than a depth threshold $\delta_D$ (set to 10mm) to handle outliers.

We also include three regularization losses to improve our model's generalization ability.
$\mathcal{L}_\text{disp} = || G_d \odot M_{G_d} ||_2$ is a regularization term on the displacement map to constrain the final points close to the surface and prevent overfitting, 
where $M_{G_d}$ is a mask to only penalize the points whose displacement values are larger than $\delta_\text{disp}$ (set to 10mm).
$\mathcal{L}_s$ is a Total Variation (TV) loss applied on the $256^2$ UV position map $G_o$ to encourage a smooth surface.
$\mathcal{L}_\text{kl}$ is the common Kullback-Leibler (KL) divergence prior applied on the latent space in VAE training.

In summary, the complete training loss is the weighted sum of these loss terms:
\begin{equation}
\setlength{\abovedisplayskip}{3pt}  
\setlength{\belowdisplayskip}{3pt}  
\begin{split}
    \mathcal{L} =& \;\lambda_\text{pho} \mathcal{L}_\text{pho} + \lambda_\text{per} \mathcal{L}_\text{per} + \lambda_d \mathcal{L}_d + \lambda_\text{rd} \mathcal{L}_\text{rd} \\
    &+ \lambda_m \mathcal{L}_m + \lambda_s \mathcal{L}_s + \lambda_\text{disp} \mathcal{L}_\text{disp} + \lambda_\text{kl} \mathcal{L}_\text{kl}
\end{split}
\label{eq: loss}
\end{equation}

\subsection{Network Structures \& Implementation Details} 
\label{sec:net}

Our network is trained in a variational auto-encoding fashion~\cite{DBLP:journals/corr/KingmaB14} following DAM and PiCA~\cite{DBLP:journals/tog/LombardiSSS18, DBLP:conf/cvpr/MaSSWLTS21, DBLP:journals/tog/LombardiSSZSS21}.
The encoder comprises 5 and 7 convolution layers (with the last 5 layers shared) respectively and encodes a UV position map, which is converted from a coarse tracked mesh ($\sim$ 5K vertices) and an average texture map into a latent code $\bm{z} \in \mathbb{R}^{8 \times 8 \times 4}$ as in PiCA~\cite{DBLP:conf/cvpr/MaSSWLTS21}.
The average texture map is obtained from an open-mouth expression by averaging the unwrapped textures of all camera views.
Note that the tracked mesh does not contain vertices for the tongue and teeth.

The decoder contains 5/7/7 convolution layers and predicts a position map, a displacement map, and a feature map, all of which are used for later radiance decoding.
The position map $\bm{G}_p$ at $1024^2$ represents a coarse (i.e., less detailed) surface as it is upsampled from $\bm{G}_o$ at $256^2$, which is supervised by the input coarse mesh $\mathcal{M}$.
The displacement map $\bm{G_d}$ at $1024^2$ increases details to compensate for lost geometric details of the coarse mesh $\mathcal{M}$ during estimation.
The $32-D$ feature map $F$ at $1024^2$ contains local appearance information around the point for radiance decoding.

We set loss weights of $\{\lambda_\text{pho}, \lambda_\text{per}, \lambda_d, \lambda_\text{rd}, \lambda_m, \lambda_s, \lambda_\text{disp},$ $\lambda_\text{kl}\}$ in Eq.~\eqref{eq: loss} as $\{5, 0.1, 0.1, 0.2, 0.2, 1, 0.1, 0.001 \}$ respectively.
And the G/E/P-stages take 10/15/5 epochs, respectively.

\section{Experiments}

We test on the Multiface dataset~\cite{DBLP:journals/corr/abs-2207-11243}.
It is an open-sourced multi-view human face dataset that captures high-quality facial details from a camera array.
Processed data include calibrated camera parameters, tracked meshes, and unwrapped UV texture maps ($1024 \times 1024$).
We obtain a depth map for each frame individually using Metashape~\cite{metashape} based on the provided camera parameters.

Following MVP~\cite{DBLP:journals/tog/LombardiSSZSS21}, we use downsampled images ($1024 \times 667$) during training.
Unless stated otherwise, all our experiments are trained on a subset of expressions and tested on held-out expressions (15 randomly chosen expressions and fixed), resulting in $\sim$11K frames for training and $\sim$1K frames for testing.
Following PiCA~\cite{DBLP:conf/cvpr/MaSSWLTS21}, MSE and LPIPS are calculated based on image pixels under the rasterized mask for evaluation.

\subsection{Comparisons with State-of-the-Art Methods}

We compare with DAM, PiCA, and MVP to demonstrate the superiority of our approach via the rendered images under novel expressions.
In Tab.~\ref{tab:view_synthesis}, {\name} achieves the best MSE (up to 5.37 lower than the $2^\text{nd}$ best).
Fig.~\ref{fig:view_synthesis} and Fig.~\ref{fig:sota_comp_s3} demonstrate that {\name} produces more realistic facial renditions, especially in challenging facial regions (e.g., eyes, beard, mouth interior).
Refer to our Supp. Mat. for video comparisons. 

\begin{table}[!t]
\centering
\resizebox{0.95\linewidth}{!}{
\begin{tabular}{l|c|c|c|r}
\toprule
\multirow{2}{*}{Methods} & Training & \multirow{2}{*}{MSE $\downarrow$} & \multirow{2}{*}{PSNR $\uparrow$} & Inference \\
& Data & & & Time (ms) \\ 
\midrule
NeRF & single frame & 13.63 & 36.79 & 38392 \\
{\name} (Ours) & single frame & 17.22 & 35.77 & 524 \\
{\name} (Ours) & 49 frames & 18.75 & 35.54 & 531 \\ 
\bottomrule
\end{tabular}
}
\caption{
\textbf{Comparison with NeRF.}
Our NPVA produces high-fidelity rendering results comparable to NeRF with a rather faster rendering speed ($\sim 70 \times$), when trained both on a single frame and multiple frames.
}
\vspace{-2mm}
\label{tab:nerf}
\end{table}

\subsection{Ablation Studies}
\label{sec: ablation}

In this section, we present a series of ablation studies to verify the effectiveness of our major design choices.

\begin{figure*}[tb]
\centering
\includegraphics[width=\textwidth]{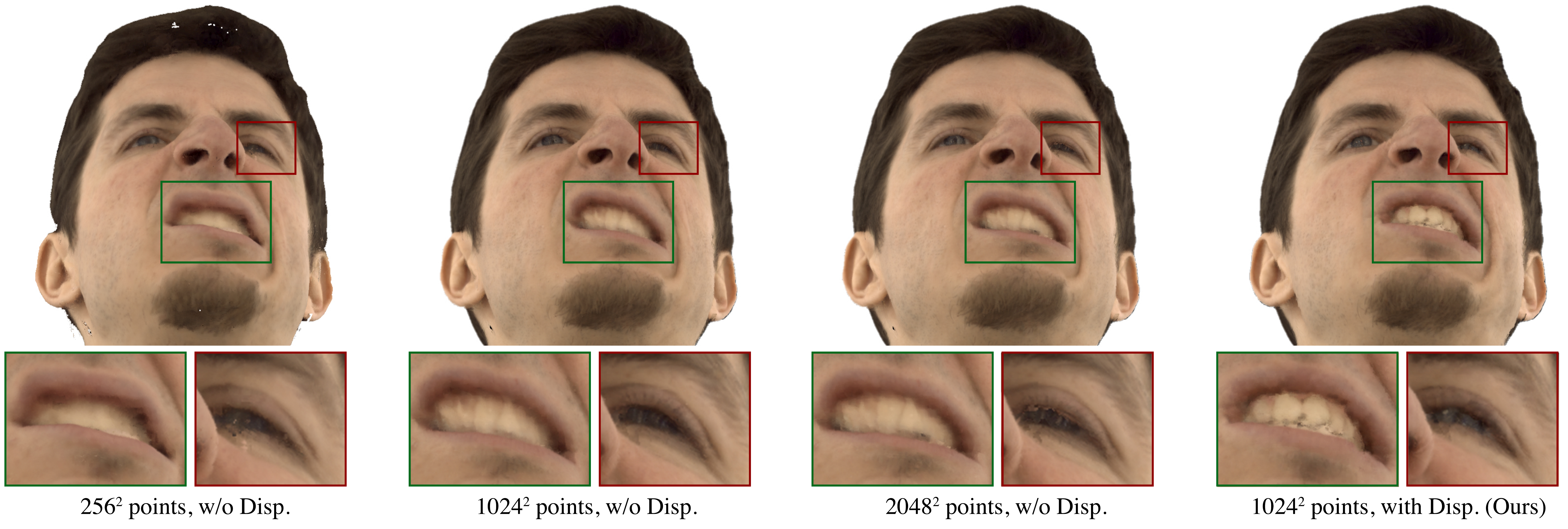}
\caption{\textbf{Ablation Study on Different Point Number and Displacements.}
When the number of points is low (i.e., $256 \times 256$), the model produces unsatisfactory results, even with holes present.
As the point number grows to $1024\times1024$, the model obtains sufficient points and fills holes, generating improved visual results.
Increasing the point number to $2048\times2048$ results in perceptually similar facial renditions.
At this time, introducing our displacements can further improve the model performance, generating more realistic results, particularly in challenging facial regions (i.e., eyes and mouth interior).
}
\label{fig:num_points}
\end{figure*}

\paragraph{Effect of using different numbers of points.} 
We investigate the impact of using different numbers of points on the rendering quality and inference time in Tab.~\ref{tab:ablation-study} (a.1)-(a.3).
The visual results are shown in Fig.~\ref{fig:num_points}.
When the point number is small (i.e., {\name}-256), {\name} generates poor results (46.45 MSE) and contains holes due to insufficient point resolution.
With increased point number ({\name}-1k), we notice obvious improvement (from 46.45 to 26.36 MSE) and do not see holes, indicating the point resolution is sufficient.
Further increasing the point number ({\name}-2k) results in perceptually very similar results and slightly worse MSE (27.36 vs. 26.36).

\begin{figure*}[tb]
\centering
\includegraphics[width=\linewidth]{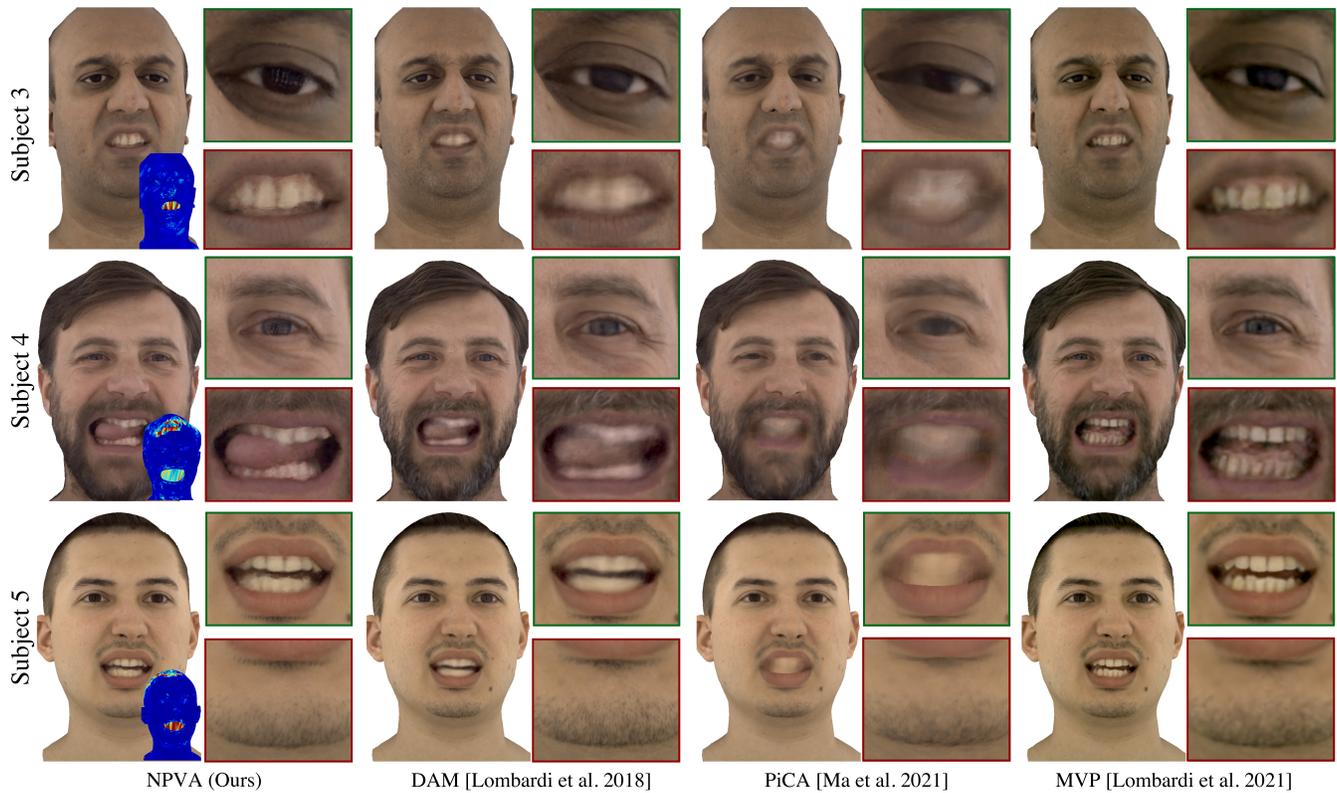}
\caption{\textbf{Qualitative Comparisons with State-of-the-Art Methods.} 
}
\label{fig:sota_comp_s3}
\end{figure*}

\paragraph{Importance of using an extra displacement map.}
Introducing an extra displacement map reduces MSE from 26.36 to 23.70 (Tab.~\ref{tab:ablation-study} (a.2)).
A visual comparison is provided in Fig.~\ref{fig:num_points}.
We can also see using a displacement map is much more effective than using more points.
This is because using a displacement map enables a more flexible arrangement of the neural points so that they can not only move on the surface (i.e., along the tangent plane) but also move along the normal direction, forming a thicker shell with increased capacity (see Fig.~\ref{fig:view_synthesis} and Fig.~\ref{fig:sota_comp_s3}).

\begin{figure}[tb]
\centering
\includegraphics[width=0.9\linewidth]{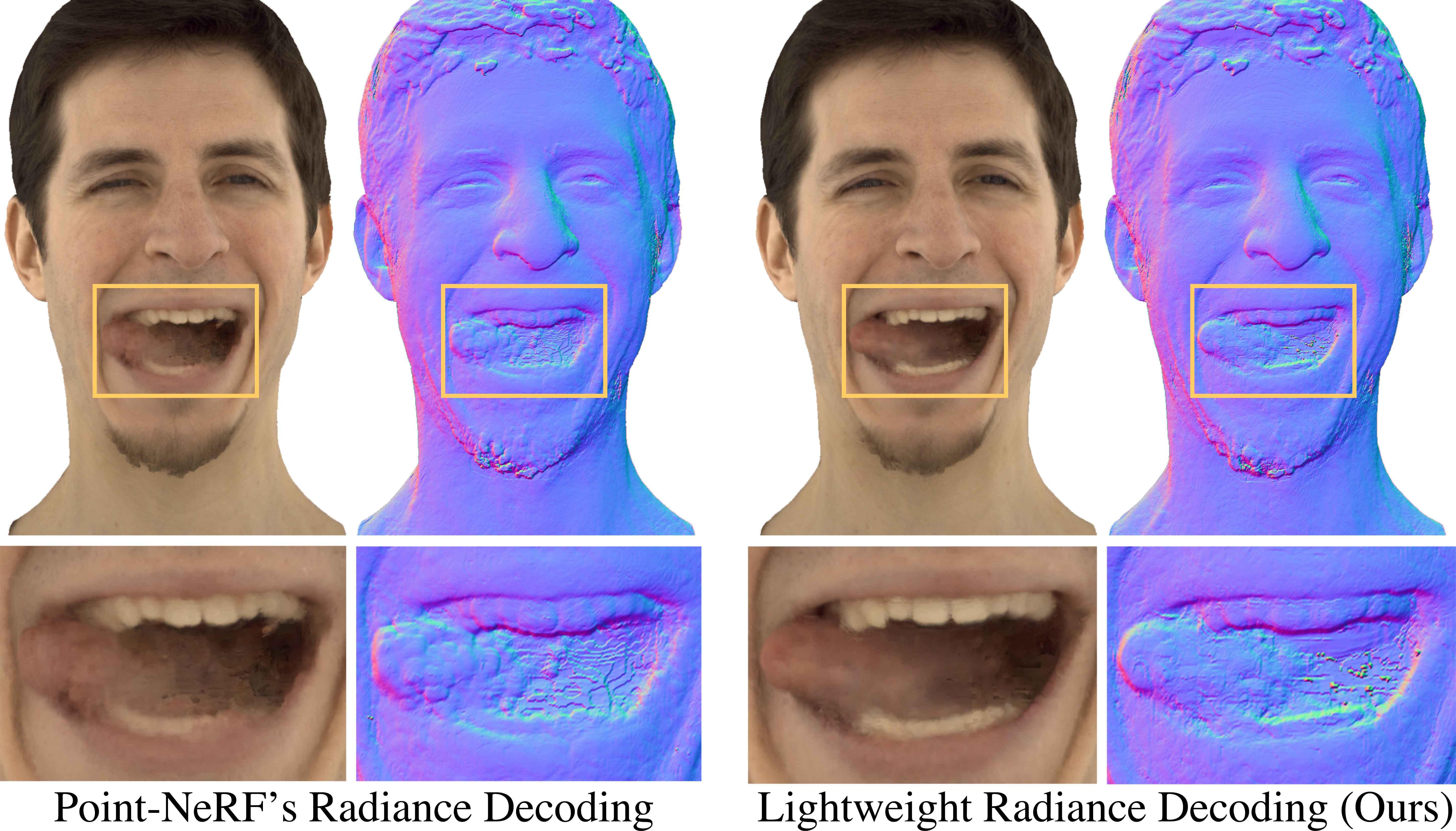}
\caption{\textbf{Effect of lightweight radiance decoding.} 
Our lightweight radiance decoding not only gains $\sim 7 \times$ speedup over Point-NeRF's radiance decoding, but also produces better facial renditions for unseen expressions.
}
\label{fig:decoding_comp}
\end{figure}

\paragraph{Influence of the lightweight radiance decoding process.} 
Using the proposed lightweight radiance decoding in Sec.~\ref{sec: NVPA Decoding} not only greatly reduces the inference time (3129 vs. 482 ms), but also improves the quality (23.70 vs. 24.51 MSE).
A visual comparison is provided in Fig.~\ref{fig:decoding_comp}.
A possible explanation is that using networks with too much capacity may cause overfitting and hinder the generalization of novel expressions for dynamic scene modeling tasks.

\begin{figure}[tb]
\centering
\includegraphics[width=0.95\linewidth]{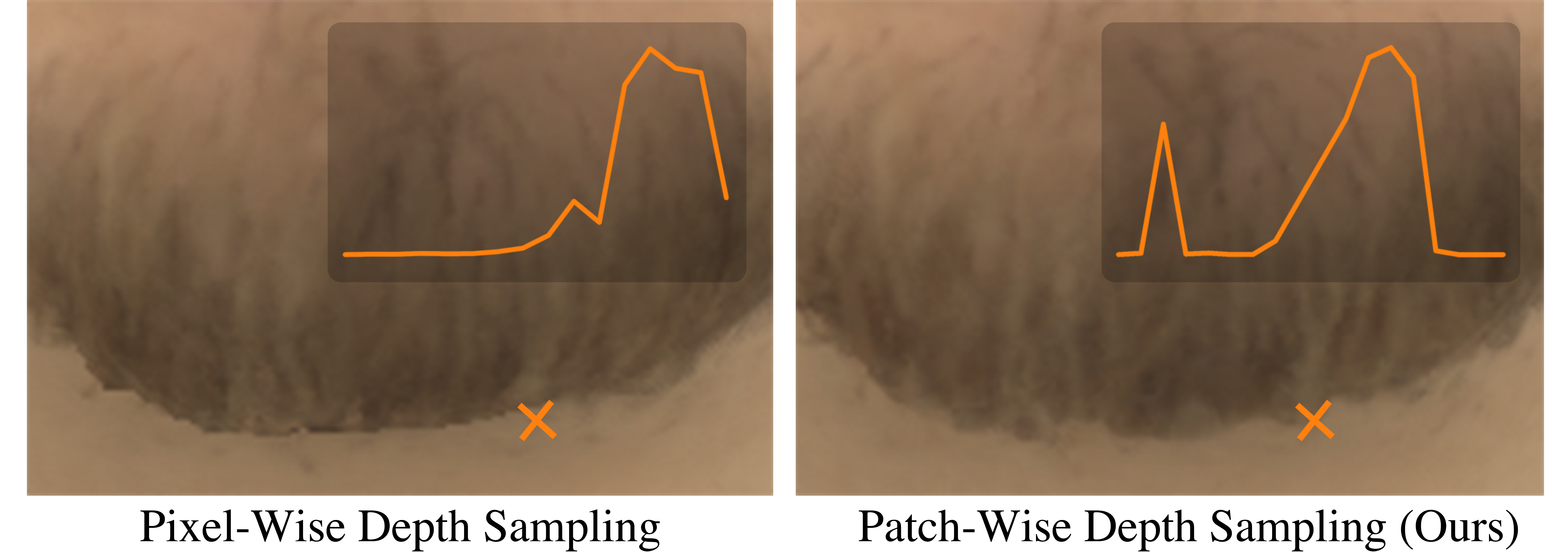}
\caption{\textbf{Effect of different depth-guided shading point sampling strategies.}
There are two depth levels around the orange mark.
With pixel-wise depth sampling~\cite{DBLP:conf/siggrapha/LinPXYSBZ22}, the model generates ``mesh-like'' artifacts.
In contrast, our patch-wise depth sampling samples around both the front and back depth levels, leading to more realistic results.
}
\label{fig:d_sampling_comp}
\end{figure}

\paragraph{Different shading point sampling methods.} 
We investigate the impact of different shading point sampling methods on a given ray.
Tab.~\ref{tab:ablation-study} (c.1, c.2) present results obtained using different sampling methods.
Our patch-wise depth-guided sampling method reduces MSE from 24.92 to 23.70 compared to pixel-wise depth-guided proposed in ~\cite{DBLP:conf/siggrapha/LinPXYSBZ22}.
Using ``Pixel-Depth'' easily causes inaccuracies, especially for the jaw region with two different depth levels, resulting in a mesh-like beard rendering (see Fig.~\ref{fig:d_sampling_comp}).
Training with a naive sampling strategy is slower and could not give a similar result (406.49 MSE) compared to the depth-guided sampling methods when using only 20 sample points per ray ($\sim$200 in NeRF).

\begin{figure*}[tb]
\centering
\includegraphics[width=\linewidth]{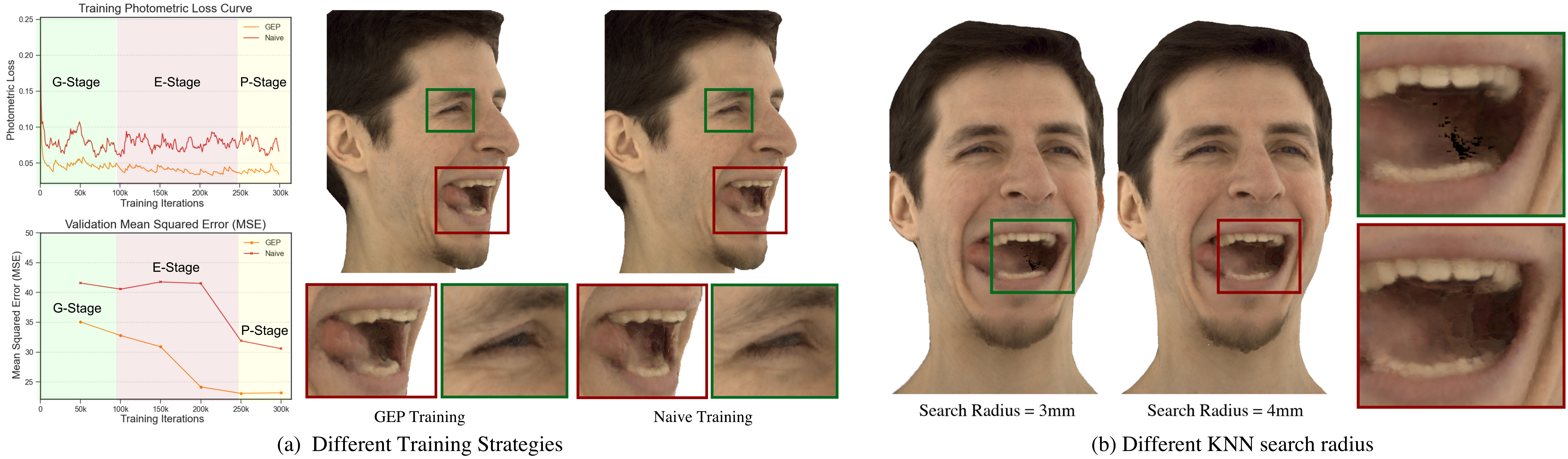}
\caption{\textbf{Comparisons on different training strategies (a) and different KNN search radius (b).}
The left part of (a) shows the photometric losses and validation MSE during training with different training strategies, while the right part of (a) shows qualitative rendering results.
The model using our GEP training strategy converges to a lower photometric loss and achieves lower validation MSE, thereby producing more realistic facial reditions.
As expected, the model using our GEP training strategy achieves a significant reduction of validation MSE during the E-stage. 
(b) shows that a larger search radius can eliminate holes in some extreme expressions with longer processing time.
}
\label{fig:training_search_r}
\end{figure*}

\paragraph{Different KNN search radius.}
Increasing the KNN search radius from 3mm to 4mm improves rendering results (23.54 vs. 23.70 MSE) with longer processing time (563 vs. 482 ms).
As shown in Fig. \ref{fig:training_search_r}b, a larger search radius can eliminate holes in some extreme expressions.

\paragraph{GEP training strategy.} 
We compare our GEP ray sampling strategy (Sec.~\ref{sec:GEP}) with a naive alternative that always uses uniformly sampled rays.
GEP achieves lower MSE (23.70 vs. 30.08) in Tab.~\ref{tab:ablation-study} (d).
The training loss curves and visual comparisons are shown in Fig.~\ref{fig:training_search_r}a.
The model trained with the naive strategy converges to a sub-optimal solution.
Although achieving satisfactory renderings in smooth facial regions (e.g., skins), it struggles to handle challenging facial regions (e.g., eyes and mouth interior).
In contrast, the model trained with our GEP strategy allocates more computing budget to these difficult regions and obtains more realistic facial renditions.

\subsection{Analysis on Volumetric Methods}

We compare with single-frame NeRF fitting~\cite{DBLP:conf/eccv/MildenhallSTBRN20}, which can be viewed as the upper limit of different volumetric avatars.
The results are shown in Tab.~\ref{tab:nerf} and Fig.~\ref{fig:NeRF_comp}.
On single-frame fitting, our NPVA generates high-fidelity results comparable to NeRF while being $\sim 70 \times$ faster (524 vs. 38392 ms) during inference.
What's more, our {\name} can handle dynamic scenes (a 49-frame sequence) effectively with minor performance drops.
In Fig.~\ref{fig:NeRF_comp}, we notice {\name} also generates visually better results (e.g., sharper and more realistic reflection effects) than NeRF, possibly due to the help of coarse geometry prior and the perceptual loss.

\section{Conclusion \& Discussion}

In this paper, we present a novel volumetric representation based on movable neural points for animatable avatar creation, focusing on both high-quality rendering and time efficiency.
To ensure controllability and accurate expression control, we guide point locations with decoded coarse meshes of target expressions and constrain the points around the surface, which is supervised with the driving signal. 
To further enhance rendering quality, we increase the point number and incorporate an additional displacement map that adaptively adjusts after training.
Moreover, our approach features three technical innovations tailored to improve training and rendering efficiency: lightweight radiance decoding, patch-wise depth-guided sampling, and a GEP training strategy.

\begin{figure*}[tb]
\centering
\includegraphics[width=\linewidth]{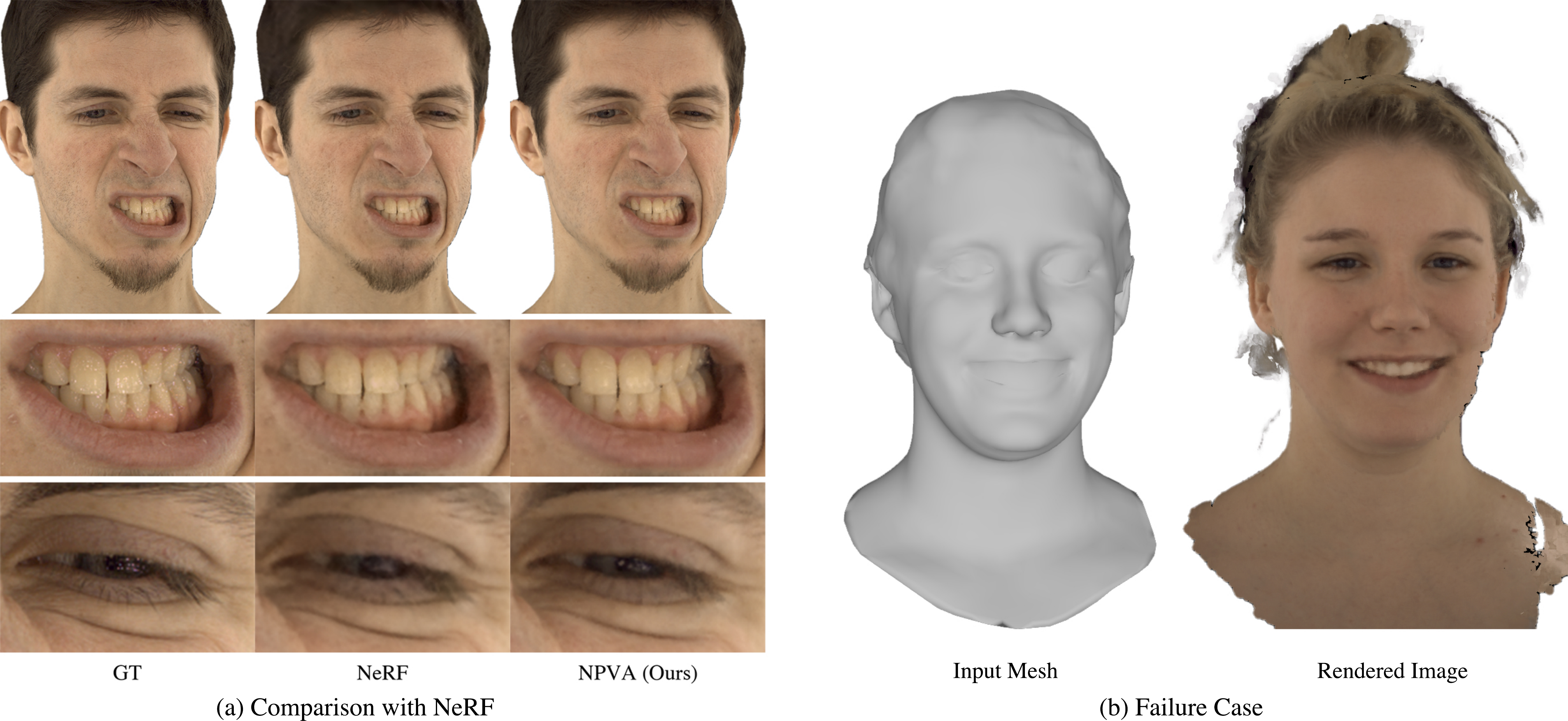}
\caption{\textbf{Qualitative comparison with NeRF and our failure case.} 
(a) shows our qualitative comparison with NeRF.
When trained on one single frame, our {\name} achieves results even visually better than NeRF (i.e., sharper and more realistic reflection effects), particularly in challenging facial regions (i.e., eyes and mouth interior) possibly due to the help of coarse geometry prior and the perceptual loss.
(b) is one failure example, in which the regions far outside the rasterized mask contain obvious artifacts. 
This result is obtained by relaxing the constraint applied on the displacement map during training. 
Since no prior knowledge about the hair is provided in the coarse mesh, we replace our depth-guided shading point sampling strategy with the NeRF sampling strategy.
}
\label{fig:NeRF_comp}
\end{figure*}

\paragraph{Limitation. }
We rely on coarse mesh tracking for modeling and optimization, which generally works well but does not account for very long hair or diverse hairstyles, such as those not present in tested female subjects. When relaxing the regularization on the displacement map for these cases, our method tends to produce blurry results for novel expressions (Fig.~\ref{fig:NeRF_comp}b).

\begin{acks}

This work was supported by the Natural Science Foundation of China (Project Number 62132012) and Tsinghua-Tencent Joint Laboratory for Internet Innovation Technology. 

\end{acks}

\bibliographystyle{ACM-Reference-Format}
\bibliography{sample-bibliography}

\clearpage

\end{document}